\documentclass[lettersize,journal]{IEEEtran}
\usepackage{amsmath,amsfonts}
\usepackage{algorithmic}
\usepackage{algorithm}
\usepackage{array}
\usepackage[caption=false,font=normalsize,labelfont=sf,textfont=sf]{subfig}
\usepackage{textcomp}
\usepackage{stfloats}
\usepackage{url}
\usepackage{verbatim}
\usepackage{graphicx}
\usepackage{cite}
\usepackage{booktabs}
\usepackage{multirow}
\usepackage{pifont}
\usepackage[hidelinks]{hyperref}
\begin{document}

\title{Skeleton2Stage: Reward-Guided Fine-Tuning for Physically Plausible Dance Generation}

\author{Jidong Jia, Youjian Zhang, Huan Fu, and Dacheng Tao,~\IEEEmembership{Fellow,~IEEE,}
\thanks{J. Jia is with School of Computer Science, Shanghai Jiao Tong University, Shanghai 200240, China (\url{jjd1123@sjtu.edu.cn}).}
\thanks{Y. Zhang is with Bosch, Shanghai 201206, China (\url{Youjian.ZHANG@cn.bosch.com}).}
\thanks{H. Fu is with Youku, Alibaba, Beijing 100124, China (\url{hufu6371@uni.sydney.edu.au}).}
\thanks{D. Tao is with Nanyang Technological University, Singapore (\url{dacheng.tao@ntu.edu.sg}).}
}

\markboth{Journal of \LaTeX\ Class Files,~Vol.~14, No.~8, August~2021}%
{Shell \MakeLowercase{\textit{et al.}}: A Sample Article Using IEEEtran.cls for IEEE Journals}

\IEEEpubid{0000--0000/00\$00.00~\copyright~2021 IEEE}

\maketitle

\begin{abstract}
Despite advances in dance generation, most methods are trained in the skeletal domain and ignore mesh-level physical constraints. As a result, motions that look plausible as joint trajectories often exhibit body self-penetration and Foot-Ground Contact (FGC) anomalies when visualized with a human body mesh, reducing the aesthetic appeal of generated dances and limiting their real-world applications. We address this skeleton-to-mesh gap by deriving physics-based rewards from the body mesh and applying Reinforcement Learning Fine-Tuning (RLFT) to steer the diffusion model toward physically plausible motion synthesis under mesh visualization. Our reward design combines (i) an imitation reward that measures a motion's general plausibility by its imitability in a physical simulator (penalizing penetration and foot skating), and (ii) a Foot-Ground Deviation (FGD) reward with test-time FGD guidance to better capture the dynamic foot-ground interaction in dance. However, we find that the physics-based rewards tend to push the model to generate freezing motions for fewer physical anomalies and better imitability. To mitigate it, we propose an anti-freezing reward to preserve motion dynamics while maintaining physical plausibility. Experiments on multiple dance datasets consistently demonstrate that our method can significantly improve the physical plausibility of generated motions, yielding more realistic and aesthetically pleasing dances. The project page is available at: \href{https://jjd1123.github.io/Skeleton2Stage/}{Skeleton2Stage Project Page}.
\end{abstract}


\section{Introduction}
\IEEEPARstart{D}{ance} is a universal art form for expressing emotions and conveying messages \cite{lamothe2019dancing, whydance1}, widely used in film production, game development, and virtual reality.

Despite the complexity of skeletal movements in human dances and their intricate relationship with the music condition, recent deep-learning-based methods \cite{li2021learn, siyao2022bailando, tseng2023edge, Luo_2024_CVPR} have made significant progress in generating high-quality, natural, and diverse dances that align well with the given music.

However, due to the complexity of body meshes' representation, most existing methods overlook skinned meshes when training with dance data. Instead, they typically represent actions through bone rotations in the skeletal form. Therefore, when the generated motions are visualized with human body meshes, many physically implausible phenomena might occur, such as body interpenetration and imbalanced movement. These issues significantly degrade the aesthetic appeal and realism of the final visual results \cite{hoyet2012push}.

To tackle the problem, we introduce physics-based rewards to incorporate physical constraints from body meshes into the generative models via RLFT. Specifically, we first train an imitation policy on expert datasets. The well-trained imitation policy can control a physically simulated character in IssacGYM \cite{makoviychuk2021isaac} to mimic the given motion. Then, we use the trained policy to construct an imitation reward that evaluates the physical plausibility of the generated motion based on its imitability. The effectiveness of this reward stems directly from the simulator's constraints, which makes the imitation policy struggle, or even fail, to replicate physically implausible motions (\textit{e.g.} body penetration, foot skating), naturally yielding a lower reward. Thereby, the imitation reward can implicitly impose the mesh-based physical constraints into generative models. Besides the imitation reward that evaluates the general physical plausibility, we further introduce an FGD reward and test-time FGD guidance to improve Foot-Ground Contact (FGC) realism, which is challenging due to the dances' dynamic foot-ground interaction.
\IEEEpubidadjcol

Furthermore, during the process of RLFT, we find that the physics-based rewards tend to favor movements with small magnitude, which encourages the generative models to generate freezing motions. This happens because freezing motions are intrinsically less prone to mesh-level artifacts (e.g., self-penetration, foot skating) and are also easier for the imitation policy to track in the simulator, thereby receiving high imitation rewards. Therefore, we propose an anti-freezing reward for models to balance the preference for freezing motions. Specifically, we evaluate the magnitude of the generated motions by computing the velocity and acceleration of pose and translation parameters. By combining the anti-freezing reward with physics-based rewards, our method can mitigate the bias towards freezing motions and encourage large movements while improving the physical plausibility. 

To validate our method, we fine-tune the state-of-the-art (SOTA) dance diffusion models EDGE \cite{tseng2023edge}, POPDG \cite{Luo_2024_CVPR}, GENMO \cite{genmo2025}, and Bailando++ \cite{10264209} and evaluate the results on the AIST++ \cite{li2021learn} and PopDanceSet \cite{Luo_2024_CVPR}. Experiments show a significant reduction in physical implausibility, such as body interpenetration and abnormal FGC. Several metrics are designed to measure these physical improvements quantitatively. Our core contributions are as follows,
\begin{itemize}
    \item{\textbf{Exposing and Bridging the ``Skeleton-to-Mesh'' Gap:}} We identify and address a critical yet often-overlooked gap between the skeleton motion generation and mesh body visualization. To bridge this, we introduce Skeleton2Stage, a novel framework that employs RLFT with a carefully designed physics-based rewards system.
    \item{\textbf{Effectively leveraging physical priors from simulators:}} Our method provides a simple yet effective way to distill physical priors from a physics simulator into generative models. We first find that an imitation policy serves as a reliable proxy for assessing the physical plausibility of motions. We then adopt it as a physics-based reward for RLFT, thereby overcoming the simulator's non-differentiability and successfully incorporating simulator-derived physical priors into generative models.
    \item{\textbf{Physics-Based Rewards System:}} We propose a multi-faceted reward system for RLFT. We first use the pretrained imitation policy as a physical-aware reward, imposing general physical constraints---especially those from skinned mesh---into the RLFT. Then we introduce the FGD reward and guidance to further rectify FGC artifacts, which is challenging due to the dynamic nature of the dance. Finally, we propose an anti-freezing reward to balance the rewards' preference for freezing motions.
    \item{\textbf{SOTA Physical Plausibility and Visual Quality:}} Our experimental results demonstrate a significant improvement in the physical plausibility of the generated dances, including penetration and foot-ground contact. The visual quality of the generated results has also been greatly enhanced in terms of realism and aesthetics.
\end{itemize}

\section{Related Work}
\subsection{Human Motion Generation and Music to Dance}
Generating realistic human motion has been extensively studied. Previous approaches \cite{10.1145/566654.566607, 10.1145/566654.566605, arikan2002interactive} primarily rely on graph-based methods.
They decompose motions into clips and recombine them according to predefined principles. 
However, these methods struggle to generate diverse human motions, especially dances that exhibit variations in speed, length, and tempos. This limitation arises from the reliance on fixed motion units and rigid composition rules. In recent years, with the emergence of deep learning and large-scale human motion datasets \cite{AMASS:ICCV:2019, li2021learn, lin2023motionx, li2023finedance, Luo_2024_CVPR,zhang2025opendance}, numerous works have explored the use of various neural networks to generate diverse human motion \cite{NEURIPS2022_40bfe617, tevet2023human,NEURIPS2023_3fbf0c1e,liang2024omgopenvocabularymotiongeneration,dai2024motionlcmrealtimecontrollablemotion, genmo2025, Xiao_2025_ICCV}. For instance, in the domain of music-to-dance generation, recent methods utilize various network structures, including CNNs \cite{10.1145/2897824.2925975}, RNNs \cite{10.1145/3240508.3240526, yalta2019weakly, alemi2017groovenet, huang2023dance}, GCNs \cite{9008261, 10.1145/3394171.3413932, ferreira2021learning}, GANs \cite{lee2019dancing, sun2020deepdance}, Transformers \cite{li2023danceformer, li2020learning, li2021learn, siyao2022bailando, 10264209,li2025souldance,wang2025dancechat,zhang2025opendance, yang2025megadance, fan2025align} and Diffusion Models \cite{tseng2023edge, alexanderson2023listen, li2024lodge, huang2024beat, zhang2024bidirectional, Luo_2024_CVPR, li2023finedance}, to better capture the intricate relationship between the joint movements and its accompanying music.

While most existing methods focus on improving the synchronization between music and human joint movements, they often overlook mesh-level constraints from the laws of physics (\textit{e.g.} human body skin collisions). Consequently, these methods tend to generate physically implausible motions when visualized with a human body mesh. In contrast, our method introduces physics-based rewards and instills physical knowledge from heuristic constraints and physical simulators into the diffusion model.
\subsection{Physics-Based Human motion modeling}
Physics-based human motion imitation is first utilized to generate realistic and controllable locomotion for characters in the physical simulator \cite{liu2017learning, yuan2020residual, liu2018learning, peng2018deepmimic, wang2017robust, merel2017learning, 10.1145/3386569.3392381, 10.1145/3355089.3356501, 10.1145/3355089.3356536}. Recent advancements have also adopted physics-based human motion imitation for more downstream tasks such as 3D human pose estimation \cite{8014743, rempe2020contacthumandynamicsmonocular, shimada2020physcapphysicallyplausiblemonocular, shimada2021neuralmonocular3dhuman, yuan20183d, yuan2019egoposeestimationforecastingrealtime, isogawa2020opticalnonlineofsightphysicsbased3d, yi2022physicalinertialposerpip, yuan2021simpoesimulatedcharactercontrol, luo2022embodiedsceneawarehumanpose, luo2022dynamicsregulatedkinematicpolicyegocentric} and 3D human motion generation \cite{yuan2023physdiff, yao2023moconvqunifiedphysicsbasedmotion, gillman2024selfcorrecting,li2025morphmotionfreephysicsoptimization}.

Among these methods, \cite{yuan2023physdiff,li2025morphmotionfreephysicsoptimization} are most related to our method. They use the imitation policy as a physically-guided motion projection module during the inference of generative models. Compared to this straightforward combination of the generative model and imitation policy, our method uses the imitation policy as a physics-based reward to fine-tune the generative model via RLFT. This allows the generative model to internalize physical constraints from the simulator and generate physically plausible motions directly. This capability \textbf{yields two key benefits}: i) Motion projection during inference may cause issues such as jittering or even imitation failure. However, directly generating results by the fine-tuned generative model can ensure the naturalness of the movements. ii) While motion projection with the simulator is time-consuming, our method can save the time of motion projection. Moreover, prior projection-based methods disable self-collision checks in the simulator to improve imitation success rate, which limits them to addressing only foot–ground penetration rather than enforcing full-body mesh constraints (e.g., body collision) required for physically plausible mesh visualizations.

Similarly, we may also utilize some optimization-based motion refining methods \cite{tiwari22posendf,he24nrdf, wang2025pamdplausibilityawaremotiondiffusion} during the inference of generative models as \cite{yuan2023physdiff}. However, similar issues persist. First, these methods are optimization-based, which will reduce efficiency during inference. Second, they heavily rely on the quality and diversity of datasets. Third, these methods focus on individual frames of poses, potentially causing abrupt motion when applied to the entire sequence. Finally, they still cannot reliably enforce full-body mesh constraints, such as self-interpenetration avoidance, which are critical for human body mesh visualizations.

\subsection{RLFT of Diffusion Models}
With the success of fine-tuning large language models (LLMs) with RL \cite{bai2022constitutional, bai2022training, lee2023rlaif, ouyang2022training}, recent research has proposed RLFT algorithms \cite{lee2023aligning, black2024training, fan2023dpok, wallace2023diffusion, rafailov2023direct} for text-to-image diffusion models. Similarly, the field of human motion generation has also begun to adopt fine-tuning approaches to improve synthesis quality and alignment \cite{han2025reindiffuse, motioncritic2025}.

In contrast to previous methods that primarily relied on data-driven or heuristic rewards, our approach uniquely incorporates a physics-based human motion imitation policy, which controls a character in the physical simulator. By leveraging RLFT, we ensure that the physical constraints (\textit{e.g.} skin collisions and gravity) of human motion are instilled into the diffusion models, offering a more robust integration of physics than previous approaches. 

\begin{figure*}[t]
    \centering
    \includegraphics[width=\linewidth]{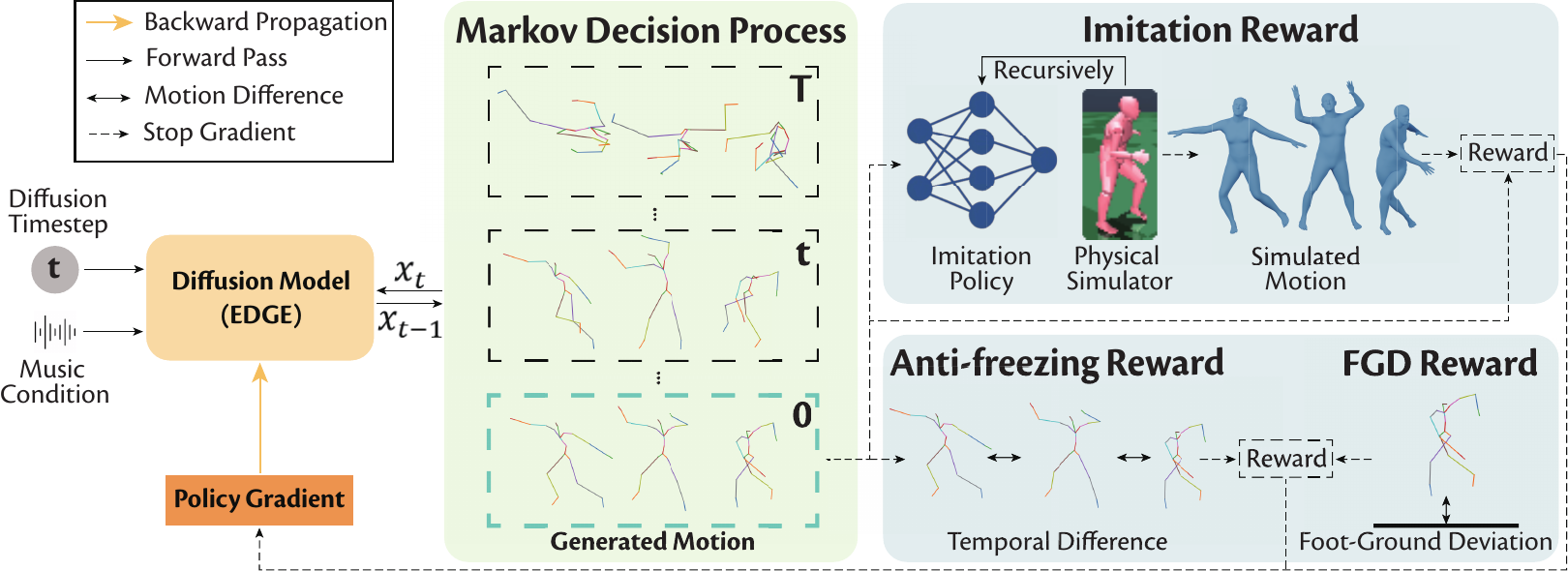}
    \caption{The overview of our method. Our method formulates the denoising process as a multi-step Markov Decision Process, allowing diffusion models to be fine-tuned via RL. To incorporate physical constraints into diffusion models, we introduce physics-based rewards, including an imitation reward assessing the general physical plausibility with an imitation policy and an FGD reward to handle the dynamic nature of dance. Additionally, we design an anti-freezing reward to mitigate the physics-based rewards' preference for freezing motions.}
    \label{fig:overview}
\end{figure*}

\section{Method}
As overviewed in Fig.~\ref{fig:overview}, our method adopts the RL strategy to instill physical constraints into dance diffusion models.
Firstly, an imitation policy that can effectively mimic the dancing sequence in the simulator is trained on expert datasets AMASS and AIST++ (in Section~\ref{sec:imitation}). This well-trained imitation policy can then act as a reward evaluator, assessing the physical plausibility---especially those aspects arising from full-body mesh constraints---of the generated motion (in Section~\ref{link}-``Imitation Reward'').
Secondly, we treat the dance diffusion denoising process as a multi-step Markov Decision (MDP) Process. This allows us to employ multiple rewards to fulfill an RLFT strategy, including the imitation reward (in Section~\ref{sec:diffusion}).

\subsection{Imitation policy for Imitation Reward}\label{sec:imitation}
To instill real-world physical constraints into our diffusion model, we need a metric to evaluate the physical plausibility of the generated motion. Inspired by \cite{yuan2023physdiff}, we leverage an imitation policy to serve this purpose. Unlike \cite{yuan2023physdiff}, we further enable the self-collision checks and utilize the trained imitation policy as a physics-aware reward evaluator to fine-tune the diffusion model with RLFT. Human motion imitation policy is commonly used in robotics to control agents in the physical simulator to replicate complex movements. Since the motions replicated in the simulator inherently satisfy the various physical laws we set for the simulator (such as gravity, object collisions, friction, etc.), we argue that the more the original motion conforms to physical laws, the more accurately it should be replicated in the physical simulator. Therefore, to utilize this characteristic during the RLFT for the motion diffusion model, we train an imitation policy to control the character with SMPL skinned mesh \cite{loper2023smpl}.

We formulate human motion imitation as an MDP \cite{yuan2020residual}, defined using states (s), actions (a), transition dynamics ($\mathcal{T}$), reward functions (r), and a discount factor ($\gamma$) as follows, 
\begin{equation}
\centering
\begin{aligned}
&s_t = \left(x_t, x_{t+1}^{res}, \psi\right),\quad s_{t+1}=\mathcal{T}\left(s_{t+1}\vert s_t,a_t,\Phi\right),\\
&a_t = x_{t+1}, \qquad\qquad \pi_{\theta}\left(a_t\vert s_t\right) = \mathcal{N}(\mu_{\theta}(s_t), \Sigma),\\
&r\left(s_t, a_t\right) = \sum_{j\in J}w_j\exp\left(-\alpha_j\left\Vert x_t^j-\overline{x}_t^j\right\Vert_2\right)\\
&\qquad\qquad\qquad\qquad+w_r\exp\left(-\alpha_r\Vert F_r\Vert_2\right),
\end{aligned}
\label{eq:imitation_mdp}
\end{equation}
where $x_t$ represents the current pose of the controlled character, including joint angles, joint velocities, rigid bodies’ positions, rotations, etc. $x_{t+1}^{res}$ is the residual between the character’s current pose and the next pose to be imitated. $\psi$ corresponds to SMPL parameters of the character. $x_t^j$ and $\overline{x}_t^j$ are the current pose sequence of imitated motions and reference motions (\textit{i.e.}, the motion given to the imitation policy to mimic). More notations will be described below (Actions, Transition, Rewards, and Training Strategy).
In each step, the imitation policy $\pi_\theta$ takes action based on the given state to control the character in the physical simulator to mimic the next pose of the given reference motion. Subsequently, the character's state transits based on the transition dynamics, and we can extract imitated motions from it. Finally, we calculate the reward in Equ.~\ref{eq:imitation_mdp} reflecting how well the reference motion is imitated. Here, we mainly illustrate Actions, Transition, and Rewards, and we recommend \cite{yuan2020residual} for a comprehensive discussion.

\textbf{Actions }
The policy outputs an action $x_{t+1}$, representing the target joint angles of the character's pose. A PD controller then drives the character towards this target pose. Additionally, to improve the imitation ability, A residual force \cite{yuan2020residual} is applied to the character's pelvis. This extra force helps to stabilize the character during movement.

\textbf{Transition }
$\mathcal{T}$ is the transition dynamics of the simulator and $\Phi$ denotes the physical constraints modeled by the physical simulator, including gravity, body collision, skeletal structure, etc. Unlike previous methods \cite{yuan2023physdiff, yao2023moconvqunifiedphysicsbasedmotion, gillman2024selfcorrecting} that utilize a humanoid as the body representation, we further embed SMPL model for a more accurate human structure, and integrate body collision handling into both the training and inference stages, Specifically, we treat each part of the SMPL model’s mesh as the collision volume (as the character in red color in Fig.~\ref{fig:overview}) and enable the collision check between most collision volumes. We further disable collision checks between the adjacent body parts, which can effectively prevent neighboring joints from being stuck.

\textbf{Rewards }
The reward has two components. The first is designed to encourage the imitated motion to match the ground truth. $J$ includes components representing human motion: local and global joint rotations, joint velocities, and 3D world joint positions. Additionally, $w_j$ and $\alpha_j$ are the weighting factors of each reward. Finally, $x_t^j$ is the t-th frame of the imitated motions represented in j, which is extracted from $s_t$, and $\overline{x}_t^j$ is the t-th frame of the ground truth in the training set. The second part is a regularization term that limits the residual force, as excessively large residual forces can harm the physical realism of the character. $\alpha_r$ and $w_r$ are weighting factors, and $F_r$ is the residual force.

\subsection{Physics-based Dance Diffusion Model}\label{sec:diffusion}
\noindent\textbf{RLFT Formulation }
We introduce an RLFT strategy to distill the physical constraints of the simulator into the diffusion model.
To conduct RLFT on diffusion models, recent works \cite{black2024training, fan2023dpok} proposed to formulate the diffusion denoising process as a multi-step MDP:
\begin{equation}
\begin{aligned}
&s_t = \left(c, T-t, x_{T-t}\right), r\left(s_t, a_t\right) = \begin{cases}
r\left(x_0, c
\right) & \text{if } t = 0, \\
0 & \text{otherwise}.
\end{cases},&\\
&a_t = x_{T-\left(t+1\right)},\ \pi\left(a_t\vert s_t\right) = p_{\theta}\left(x_{T-\left(t+1\right)}\vert c, T-t, x_{T-t}\right),&\\&r\left(x_0, c\right) = r_{imit}(x_0) + r_{anti}(x_0) + r_{FGD}(x_0),&
\end{aligned}
\label{eq:formulation}
\end{equation}
where $t$ denotes the $t \text{-th}$ decision step, while $T$ is the total number of steps. $c$ is the condition signal (\textit{i.e.} music sequence in our task). $x_{T-t}$ is the whole sequence of the denoised motion at denoising step $T-t$. $p_\theta$ is the diffusion model being fine-tuned. The reward consists of three components: $r_{imit}$, which evaluates the physical plausibility of the generated motion, $r_{anti}$, which is designed to eliminate freezing issues, and $r_{FGD}$, which aims to improve FGC realism. As our method involves two MDP formulations, some notation may be confusing. It is noteworthy that all the notations from Euq.~\ref{eq:imitation_mdp} and Euq.~\ref{eq:formulation} are independent and should be interpreted separately. For example, $t$ in Euq.~\ref{eq:imitation_mdp} denotes the time step of the MDP, while $t$ in Euq.~\ref{eq:formulation} denotes the time step of the denoising process. 

With this formulation, we can use any policy-based RL algorithm, such as REINFORCE \cite{williams1992simple}, to optimize diffusion models based on task-oriented rewards. To ensure training stability, we adopt a pure on-policy training strategy. Unlike off-policy training, which may leverage the data collected by older policies for better sample efficiency, we only use the data collected by the newest policy to update itself. Meanwhile, to enhance the numerical stability and convergence, we also normalize the reward to have zero mean and unit variance as in \cite{black2024training}. The reward's mean and standard deviation statistics are tracked for each music independently: 
\begin{equation}
A_i(x_0, c) = \frac{r_i(x_0, c)-\mu(c)}{\sigma(c)}\quad i\in\{imit,anti,FGD\}.
\end{equation}
After collecting enough trajectories, we can update the diffusion model using the policy gradient,
\begin{equation}
\begin{aligned}
\bigtriangledown_\theta J = E\Bigg[\sum_{t=0}^{T}\bigtriangledown_\theta\log p_\theta \left(x_{T-\left(t+1\right)}\vert c,x_{T-t} \right)\\*\left(\alpha A_{\text{imit}}\left(x_0,c\right)+\beta A_{\text{anti}}\left(x_0,c\right)+\gamma A_{\text{FGD}}\left(x_0,c\right)\right)\Bigg]
\end{aligned},
\end{equation}
where $p_\theta$ denotes the diffusion model, and $A_{imit}(x_0,c)$, $A_{anti}(x_0,c)$, and $A_{FGD}(x_0,c)$ are the normalized imitation, anti-freezing, and FGD rewards for guiding the optimization, with $\alpha$, $\beta$, and $\gamma$ as their weights.

\begin{table*}[t]
    \caption{Quantitative results on AIST++ dataset}
    \label{tab:performance}
    \small
    \resizebox{\linewidth}{!}{
    \begin{tabular}{lcccccccc}
    \toprule
        Method&Overall&Physical&$\text{FID}_\text{k}$ /$\text{FID}_\text{g}$ $\downarrow$&Pen. Rate $\downarrow$&PFC $\downarrow$&FGD $\downarrow$&BAS $\uparrow$&$\text{Div}_\text{k}$ /$\text{Div}_\text{g}$ $\rightarrow$\\
        \midrule
Ours&/&/&63.59/24.78&\textbf{90.38}&\textbf{0.3361}&\textbf{2.5153}&0.2991&1.95/4.16\\
         $\text{EDGE}^\diamondsuit$ (CVPR'23)&70\% &77.5\% &61.67/20.77&176.44&0.8715&11.8547&0.2847&2.03/3.79\\
         $\text{PAMD}^\heartsuit$ (arXiv'25)&62.5\%&67.5\%&56.90/19.87&142.49&0.9557&14.6663&0.3026&2.50/3.82\\
         \midrule
         $\text{FACT}^\diamondsuit$ (ICCV'21)&77.5\% &75\% &69.04/19.63&97.98$^\ast$&1.2125$^\ast$&19.8958$^\ast$&0.2380&3.07/6.87\\
         $\text{Bailando}^\diamondsuit$ (CVPR'22)&85\% &87.5\% &26.99/10.25&176.36&1.5466&50.5418&0.2320&5.83/7.30\\
         $\text{Bailando++}^\diamondsuit$ (TPAMI'23)&80\%&67.5\%&21.71/9.66&130.10&1.9124&49.7011&0.2383&6.63/7.01\\
         $\text{BADM}^\clubsuit$ (CVPR'24)&-&-&-&-&1.424&-&0.2366&-\\
         $\text{Beat-it}^\clubsuit$ (ECCV'24)&-&-&-&-&0.966&-&\textbf{0.661}&-\\
         $\text{DanceBA}^\diamondsuit$ (ICCV'25)&80\%&70\%&31.90/16.22&156.22&1.3845&48.3995&0.2506&6.69/8.13\\
         $\text{DanceChat}^\clubsuit$ (arXiv'25)&-&-&-&-&0.828&-&0.27&-\\
         $\text{OpenDanceNet}^\clubsuit$ (arXiv'25)&-&-&30.09/9.75&-&1.003&-&0.2545&-\\
         $\text{Megadance}^\clubsuit$ (NeurIPS'25)&-&-&25.89/12.62&-&-&-&0.238&5.84/6.23\\ 
         \midrule
         Ground Truth& 47.5\%& 42.5\%&19.47/9.33&135.27&1.4699&4.9451&0.2292&8.72/7.77\\
         \bottomrule
    \end{tabular}}

    \vspace{0.5ex}
    \footnotesize
    The ``Overall'' and ``Physical'' columns represent the win rate of our method over the others, \textit{i.e.} a higher value indicates better performance of our method. ``*'' means noting that FACT's good Pen. Rate, PFC, and FGD mainly come from its tendency to generate freezing motions, which is further proved by the magnitude of motion in Table~\ref{tab:ablation_freezing}. ``$\diamondsuit$'' means directly using the official checkpoint to reproduce the result. ``$\heartsuit$'' means using the official code to reproduce the result, as the official checkpoint is unavailable or broken. ``$\clubsuit$'' means directly using the result in their paper, as the official code is unavailable or unusable. Our findings, along with recent advances \cite{tseng2023edge, 10264209, Luo_2024_CVPR, huang2024beat, zhang2024bidirectional, wang2025dancechat, motioncritic2025, lin2025quest,zhang2025opendance}, suggest that FID is not a reliable metric for evaluating the realism and perceptual quality of dance generation
    (Referring to Section~\ref{sec:fid_discussion_maintext} for a detailed discussion).
\end{table*}

\noindent\textbf{Imitation Reward }\label{link}
As aforementioned, we leverage the learned imitation policy as an evaluator to determine whether the generated motion of the diffusion model is physically plausible. In this section, we just use the inference process of the trained imitation policy. 
Specifically, starting from the first pose of a motion sequence, we repeatedly compute the transition function $\mathcal{T}$ (Equ.~\ref{eq:imitation_mdp}) to obtain the next frame of imitation, eventually generating the entire sequence of imitated motion. Ideally, if the generated motion obeys the real-world physical constraints, the imitated motion should be the exact same sequence as the generated motion. However, if the generated motion somehow violates the physical laws---for instance, an arm penetrating through the body---the imitation policy will attempt to control the character to mimic the generated motion under the physical constraints of the physical simulator and produce an imitated motion that closely approximates the input motion while eliminating the interpenetration. In this case, a tracking error inevitably arises between the generated motion and the imitated motion. This error is a direct metric for physical implausibility: the more a generated motion defies physical laws, the harder it is for the policy to imitate, leading to a larger tracking error. Consequently, we measure this error and assign larger rewards to motions with smaller tracking errors. The reward is formulated as follows:

\begin{equation}
    r_{imit}\left(x_0\right) = \sum_{j\in J}w_j\exp\left(-\alpha_j\left\Vert x_0^j-\hat{x}_0^j\right\Vert_2\right),
    \label{eq:reward}
\end{equation}
where $J$ is a set in which each component is a specified representation of human motion, including local and global joint rotations, joint velocities, and 3D world joint positions. $x_0^j$ is the generated motions represented in $j$, and $\hat{x}_0^j$ is the imitated motions. Additionally, $w_j$ and $\alpha_j$ are the weighting factors of each reward, and these values are kept the same as the training process of the imitation policy to ensure consistency in evaluating the difference between the reference and imitated motion.

\noindent\textbf{Anti-freezing Reward }
The imitation reward is designed to rate higher rewards for physically plausible motions, however, we find that freezing/slow-speed motions can also receive high rewards, as they intrinsically have less physical artifact and are relatively easier to imitate. Therefore, without an anti-freezing reward, the diffusion model tends to increase the probability of generating freezing movements to maximize the reward during the RLFT process. To mitigate this bias, we propose an anti-freezing reward to encourage the generative model to generate more dynamic motions while maintaining the physical plausibility. Specifically, we compute the velocity and acceleration of the motion from the pose sequence, and apply the mean square value as the anti-freezing reward:
\begin{equation}
r_{\text{anti}}(x_0) = \overline{v(x_0)^2} + \overline{a(x_0)^2}.
\end{equation}

\noindent\textbf{Foot-Ground Deviation Reward }
This reward is designed to improve the foot-ground contact realism more accurately, including foot-ground penetration and floating. Specifically, the reward is designed as the distance between the lowest joint of the motion and the ground:
\begin{equation}
r_{\text{FGD}}(x_0) = \exp\left(-\alpha_{FGD}\left\Vert h_{foot}-h_{ground}\right\Vert_2\right).
\end{equation}

\noindent\textbf{Foot-Ground Deviation Guidance }
Inspired by \cite{dhariwal2021diffusionmodelsbeatgans}, we proposed an FGD Guidance to further enhance the FGC realism of the generated dance. We first calculate the distance between the lowest joint of the motion and the ground as $G(\mu_t,c)=\left\Vert h_{foot}-h_{ground}\right\Vert_2$. And then use its gradient to guide the predicted mean at the denoising step t. Notably, we apply this guidance based on the model's predicted contact labels, disabling it during non-contact phases, which effectively reduces foot–ground penetration and floating while preserving jump motions. Finally, we apply a post-processing step that utilizes the model's predicted contact labels. This ensures that feet remain stationary on the ground during contact phases, effectively eliminating foot-skating artifacts.

\section{Experiments}
\begin{figure*}[t]
    \centering
    \includegraphics[width=\linewidth]{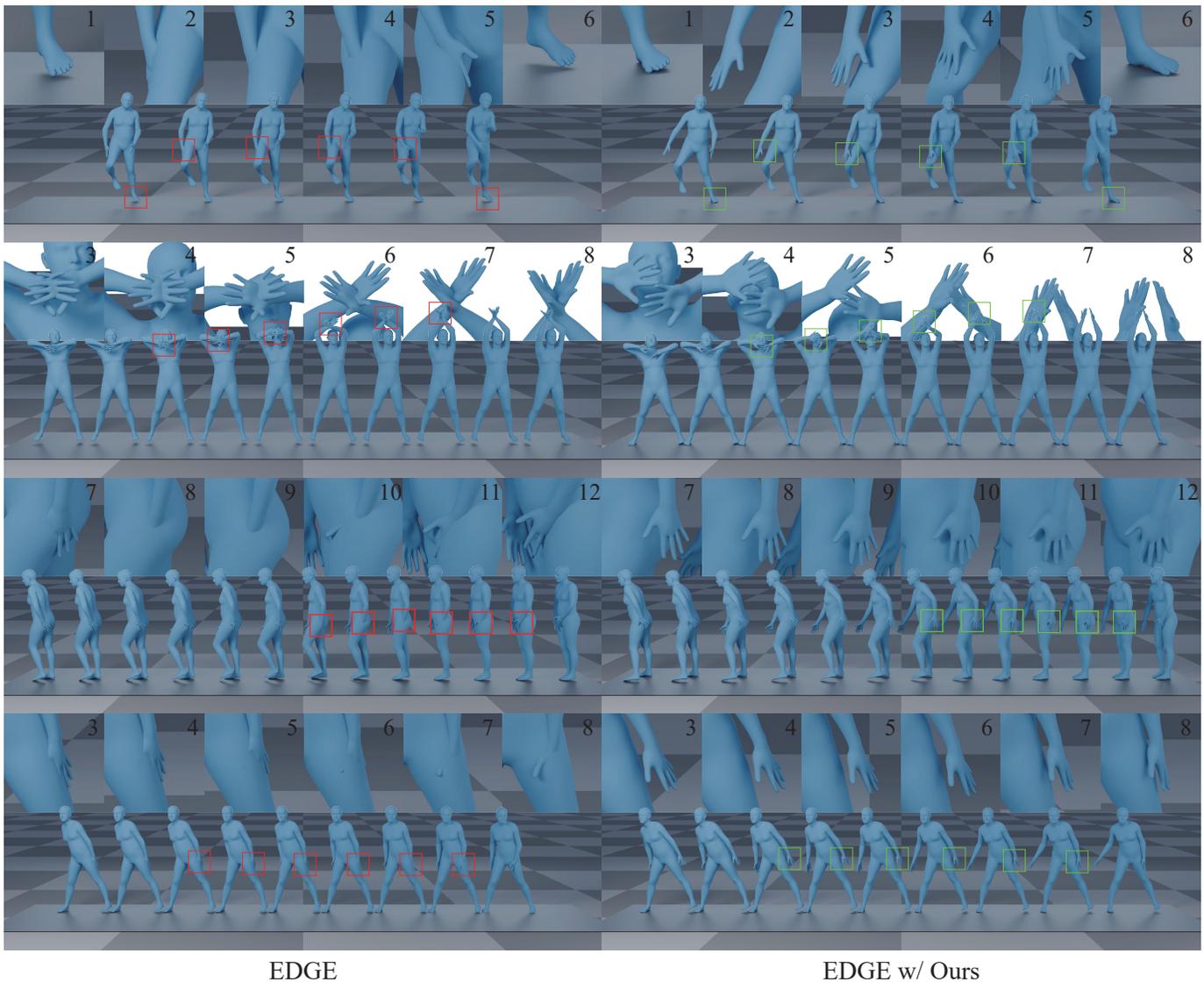}
    \caption{The visual comparisons of EDGE~\cite{tseng2023edge} and our generated motions. Both motion sequences are generated with the same music and seed. Some body parts are enlarged for a better view. The red box signifies the presence of body penetration, while the green box indicates the improvement after the RLFT. The subscript number denotes the frame number.}
    \label{fig:collision}
\end{figure*}
\paragraph{Dataset}
We conduct experiments on  AIST++~\cite{li2021learn}, which is the most widely used benchmark for dance motion generation. It contains 1,408 high-quality dance motions paired with music across diverse genres. We further evaluate our method on PopDanceSet~\cite{Luo_2024_CVPR}, which includes 736 highly dynamic dance sequences. Following EDGE~\cite{tseng2023edge}, all training samples are clipped to 5 seconds at 30 FPS.
\paragraph{Implement Details}
For the imitation policy, we adopt Isaac Gym~\cite{makoviychuk2021isaac} as the physical simulator, in which we can detect the collision between the character's torsos. The weighting factors for calculating the reward ($w_j$ and $\alpha_j$) are set to (0.6, 0.1, 0.2, 0.1) and (60, 0.2, 100, 40), respectively. And $w_r$ and $\alpha_r$ are 0.1 and 30.
For the diffusion model, we test our RLFT method on the EDGE \cite{tseng2023edge}, POPDG \cite{Luo_2024_CVPR}, and GENMO \cite{genmo2025} denoiser, and we employ denoising diffusion implicit models (DDIM) as proposed in~\cite{song2020denoising} with 50 diffusion steps and classifier-free guidance~\cite{ho2022classifier}. In each fine-tuning iteration, we sample 2,048 motions from the pretrained diffusion models, conditioned on the music in the AIST++~\cite{li2021learn} or PopDanceSet~\cite{Luo_2024_CVPR} training dataset. We accumulate gradients across 50 denoising steps of all samples and perform one gradient update. Our optimizer adopts the Adam~\cite{kingma2017adammethodstochasticoptimization} optimizer, with a learning rate set to 1e-6.
\paragraph{Evaluation Metrics}
Similar to~\cite{lin2025quest}, we evaluate our physically plausible dance generation results from three key perspectives: aesthetic quality, physical plausibility, and motion-condition consistency. For aesthetic quality, we conduct a user study (``Overall'' in Table~\ref{tab:performance}). For motion-condition consistency, we adopt the Beat Alignment Score (BAS)~\cite{siyao2022bailando}. Regarding physical plausibility, we employ both subjective and objective metrics, including a user study (``Physical'' in Table~\ref{tab:performance}), Penetration Rate (Pen. Rate), Physical Foot Contact (PFC)~\cite{tseng2023edge}, Foot-Ground Deviation (FGD), and Motion Magnitude (Mot. Mag.).

\subsection{Benchmark Performance}
We compare our method with previous dance generation approaches, as summarized in Table~\ref{tab:performance}. For fairness, all evaluation metrics are computed on 20-second dance clips.

\begin{figure*}[t]
    \centering
    \includegraphics[width=\linewidth]{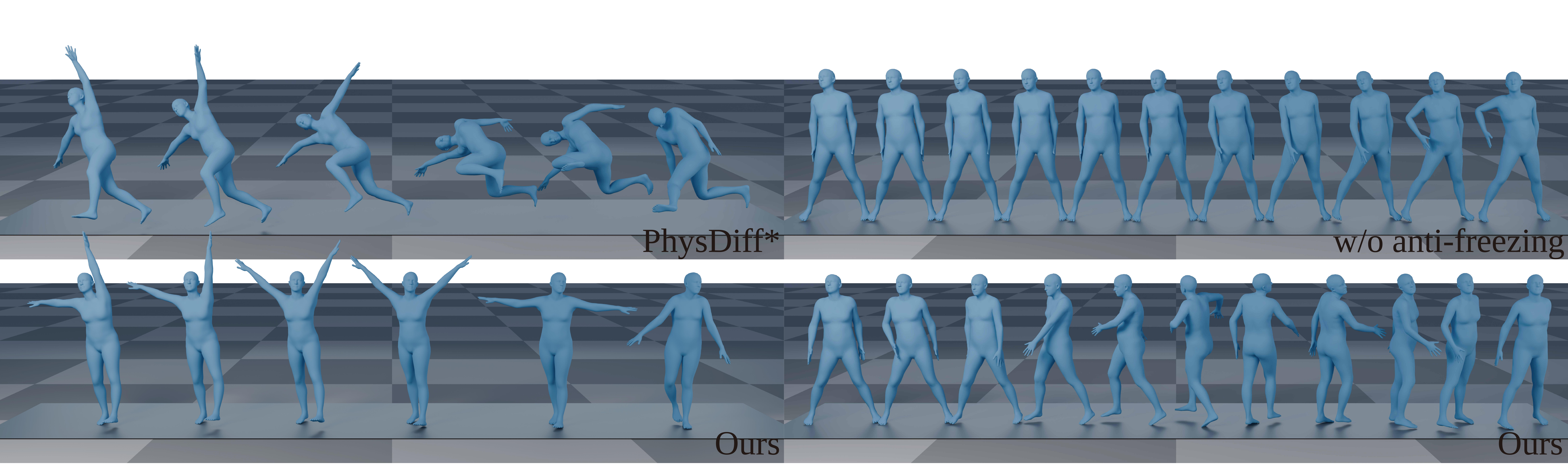}
    \caption{
    The visual comparison for ablation studies. Each compared dance pair is generated from the same audio track. Left compares the results of PhysDiff with those of our proposed method. As shown, motion projection can result in falling motions due to the inability to accurately imitate the physically implausible movements. The right presents an example result of the model trained without an anti-freezing reward, in which the model tends to generate small-amplitude movements.}
    \label{fig:ablation}
\end{figure*}

\noindent \textbf{Human Perception Results }
``Overall'' and ``Physical'' (Table~\ref{tab:performance}) denote human-perceived aesthetic quality and physical plausibility, measured as the win rate of our method over others on two questions respectively: (1) Which dance looked and felt better overall? and (2) Which dance appears more physically plausible? The study was conducted on Prolific with 40 participants under the same criteria as~\cite{tseng2023edge}. Skeleton2Stage surpasses other approaches, demonstrating a clear improvement in physical plausibility. It is worth noting that our method in Table~\ref{tab:performance} leverages the pretrained EDGE model. This means the generated dance sequences share similar motion patterns with EDGE, as shown in Fig.~\ref{fig:collision} and the supplementary videos. Nevertheless, human evaluators perceive our results as having higher overall aesthetic quality than EDGE, indicating that physical plausibility can substantially influence aesthetic perception. 

\noindent\textbf{Penetration Rate }
Pen. Rate reflects the physical plausibility of generated motions in terms of body-part penetration. As shown in Table~\ref{tab:performance}, our proposed method achieves a substantial improvement in this metric. Compared to the baseline EDGE, we reduce the penetration rate by 49\%. Although FACT also yields a low penetration rate, this is primarily due to its tendency to generate freezing motions (as discussed in Table~\ref{tab:ablation_freezing}), which are aesthetically undesirable. Qualitative comparisons between EDGE and our method are provided in Fig.~\ref{fig:collision}, where we can clearly observe that, after RLFT, our model replaces previously implausible motions with physically valid alternatives. For instance, in the first, third, and fourth rows of Fig.~\ref{fig:collision}, the model adjusts arm movements outward to avoid interpenetration; in the second row, it learns to prevent the penetration between two hands.

\noindent\textbf{Foot-Ground Contact Score }
We evaluate both the Physical Foot Contact (PFC)~\cite{tseng2023edge} and Foot-Ground Deviation (FGD) scores. The former measures the overall plausibility of foot–ground contact, while the latter is specifically designed to quantify foot–ground penetration and floating. As shown in Table~\ref{tab:performance}, our method significantly outperforms previous works, demonstrating that our approach effectively enhances the physical realism of foot–ground interactions. The PFC metric is grounded in the principle that the character in the physical simulator should obey Newton’s laws of motion, which form the theoretical basis of its definition.

\noindent\textbf{Other Non-physical Metrics }
We report the results of other non-physical metrics in Table~\ref{tab:performance} to demonstrate that our method does not deteriorate in aesthetic performance, such as beat alignment score (BAS) and diversity ($\text{Div}_\text{k}$, $\text{Div}_\text{g}$). Since our approach is not specifically designed to address beat alignment or diversity, it achieves results comparable to EDGE. It seems that diffusion-based methods tend to yield lower diversity scores when generating long dance sequences. We attribute this to the way diffusion models generate long sequences—by stitching together multiple short clips. Diversity is then computed as the average across these clips; thus, while each short clip may exhibit high diversity, the overall composition can appear less diverse.

\subsection{Ablation Study}\label{ablation}
\noindent \textbf{RLFT vs. Motion Projection }
While we use the imitation policy as an imitation reward to fine-tune diffusion models, it can also be applied as a motion projection (PhysDiff~\cite{yuan2023physdiff}) during the denoising process. We compare our method and PhysDiff, both built upon EDGE.

In our preliminary experiments, we observed that applying the physics-based projection consecutively four times (PhysDiff*(4step)) at the end of the diffusion process (the original setting in~\cite{yuan2023physdiff}) amplifies the errors caused by physically implausible motions, making imitation more prone to failure, as shown in Fig.~\ref{fig:ablation}(left) and Table~\ref{tab:ablation_proj}. Due to the extensive body interactions in dance movements, this issue occurs more frequently than in text-conditioned motion generation. Therefore, we adopt a setting that performs the physics-based projection once to achieve better stability, which we denote as ``PhysDiff*(1step)''. Nevertheless, imitation failures still appear in some cases---\textit{e.g.}, one leg passes through the supporting leg---leading to chaotic outcomes such as slipping or falling (see Fig.~\ref{fig:ablation}(left)). We infer that the performance of PhysDiff strongly depends on the success of motion imitation; however, certain penetrated motions cannot be faithfully imitated, resulting in severe failures.

\begin{table}[t]
    \caption{RLFT vs. Motion Projection}
    \label{tab:ablation_proj}
    \small
    \centering
    \resizebox{\linewidth}{!}{\begin{tabular}{lcccc}
    \toprule
        Method &Pen. Rate $\downarrow$&PFC $\downarrow$&FGD $\downarrow$&Succ. Rate$\uparrow$\\
        \midrule
         EDGE&176.44&0.8715&11.8547&100\%\\
         PhysDiff*(1step)&115.65&1.6240&11.7289&95\%\\
         PhysDiff*(4step)&\textbf{88.81}&3.7823&34.6713&45\%\\
         Ours&\underline{90.38}&\textbf{0.3361}&\textbf{2.5153}&\textbf{100\%}\\
         \bottomrule
    \end{tabular}}
    
    \vspace{0.5ex}
    \footnotesize
    \raggedright
* indicates that we reproduced PhysDiff, as the official code is not available. Key differences between RLFT and motion projection (proposed in PhysDiff) are discussed in the supplementary materials.
\end{table}

The quantitative results in Table~\ref{tab:ablation_proj} further confirm that although the penetration rate of PhysDiff*(1step) decreases due to collision checking in the physical simulator, its PFC metric deteriorates significantly. This may be attributed to abnormal foot movements: (i) in imitation failure cases, the feet lose contact with the ground; and (ii) even without falling, when the generated motions exhibit foot sliding, the imitation policy still tends to keep tracking it, which drags the feet and induces jittery stick–slip behavior due to friction in the physical simulator. In contrast, our method remains stable by assigning very low rewards to such ``failing" and foot-sliding scenarios. Moreover, unlike direct imitation, which merely reproduces similar motion patterns, our approach gradually learns to avoid these challenging cases by replacing them with physically plausible motions that are, by nature, more imitatable by the physics-based imitation policy---a key advantage introduced by RLFT. Additional qualitative results are provided in the supplementary materials.

\begin{table*}[tbp]
    \caption{The validation of the generality of our method}
    \label{tab:generality}
    \small
     \resizebox{\linewidth}{!}{
     \begin{tabular}{lcccccccc}
    \toprule
        Method&Venue&Dataset&Model&Pen. Rate $\downarrow$&PFC $\downarrow$&FGD $\downarrow$&BAS $\uparrow$&$\text{Div}_\text{k}$ /$\text{Div}_\text{g}$ $\rightarrow$\\
        \midrule
         POPDG&\multirow{2}{*}{CVPR 2024}&\multirow{2}{*}{PopDanceSet}&\multirow{2}{*}{Diffusion}&245.08&1.9265&17.7154&0.2410&4.64/6.82\\
         POPDG w/ Ours&&&&\textbf{119.41}&\textbf{1.2445}&\textbf{1.1983}&\textbf{0.2554}&4.95/6.30\\
         \midrule         
         Bailando++ w/o RL&\multirow{3}{*}{TPAMI 2023}&\multirow{3}{*}{AIST++}&\multirow{3}{*}{VQVAE+GPT}&136.62&1.5474&46.4480&0.2204&7.49/6.78\\
         Bailando++ w/ BA&&&&130.10&1.9124&49.7011&0.2383&6.63/7.01\\
         Bailando++ w/ Ours&&&&\textbf{112.49}&\textbf{1.2722}&\textbf{34.2014}&\textbf{0.2468}&7.56/7.08\\
         \midrule
         GENMO*&\multirow{2}{*}{ICCV 2025}&\multirow{2}{*}{AIST++}&\multirow{2}{*}{Diffusion}&228.04&0.5944&34.9455&\textbf{0.2355}&3.08/4.47\\         
         GENMO* w/ Ours&&&&\textbf{138.78}&\textbf{0.1703}&\textbf{5.2832}&0.2309&2.81/4.04\\
         \bottomrule
    \end{tabular}
    }

    \vspace{0.5ex}
    \footnotesize
    * indicates that we reproduce the results based on their paper, as they didn't release the correct checkpoint, training, and evaluation guidance (see Section~\ref{sec:genmo_maintext}). To ensure a fair comparison on Bailando++, we use its original RL method, modifying only the reward function. The baseline, termed ``bailando++ w/o RL'', is the pretrained GPT model from their work before RL finetuning. Then we evaluate on two different reward configurations: (1) ``Bailando++ w/ BA'' is finetuned by a beat alignment (BA) reward, the same as in their paper; (2) ``Bailando++ w/ Ours'' is finetuned by our proposed rewards.
\end{table*}

\noindent \textbf{Other Physical Rewards } Here, we do ablation studies to validate the effectiveness of other rewards.

\emph{Anti-freezing Reward:} As mentioned earlier, physics-based rewards, while effective at addressing physical issues, often drive the model to produce freezing motions, since static poses inherently avoid physical violations. To mitigate this, we develop an Anti-Freezing (AF) reward. We evaluate its effectiveness using the motion magnitude (Mot. Mag.), defined as the average temporal difference of poses across the entire sequence. This metric reflects the degree of motion freezing in the generated sequences and should not be too low, as an excessively small value indicates freezing behavior. As reported in Table~\ref{tab:ablation_freezing}, Mot. Mag. reduces significantly without anti-freezing reward (\textit{i.e.} Ours w/o AF). Fig.~\ref{fig:ablation}(right) also shows the same conclusion that it tends to generate freezing motions without anti-freezing reward, which is undesirable in practice.
\begin{table}[t]
\caption{Anti-freezing reward ablation.}
\label{tab:ablation_freezing}
\centering
\footnotesize
\setlength{\tabcolsep}{8pt}
\begin{tabular}{lc}
\toprule
Method & Mot. Mag. $\uparrow$ \\
\midrule
EDGE        & 0.4744 \\
FACT        & 0.2148 \\
Ours w/o AF & 0.3362 \\
Ours        & 0.4670 \\
\midrule
Ground Truth & 0.4881 \\
\bottomrule
\end{tabular}
\end{table}

\emph{FGD Reward\&Guidance:} We conduct ablation studies to validate the effectiveness of the FGD reward and guidance. As shown in Table~\ref{tab:ablation_FGC}, our method produces more accurate foot–ground contact with FDG reward and guidance.

\noindent\textbf{Generalization } 
We further apply our method to several other works, including POPDG~\cite{Luo_2024_CVPR}, GENMO~\cite{genmo2025}, and Bailando++~\cite{10264209}. For POPDG, we evaluate on their proposed dataset, which they indicate contains more complex dance motions. For the other two methods, we evaluate on the AIST++ dataset. As reported in Table~\ref{tab:generality}, integrating our method leads to consistent improvements across all cases. It demonstrates the generalization capability of our method.

\section{Conclusions and Discussions}
In this paper, we present Skeleton2Stage, a novel framework that addresses the physical implausibility issue caused by the skeleton-to-mesh gap in dance generation methods through RLFT. Specifically, we design a physics-based rewards system to instill physical laws derived from both physical simulators and heuristic constraints(\textit{e.g.} body collisions, gravity, and friction) into the pretrained diffusion model via RLFT. The system comprises three key components: (1) an imitation reward evaluating general physical plausibility with an imitation policy, (2) an FGD reward and guidance to handle complex foot-ground contact in dances, and (3) an anti-freezing reward mitigating models' tendency to generate freezing motions. 
Extensive experiments demonstrate that our method significantly improves the physical plausibility of generated motions across multiple metrics, particularly in reducing body penetration and enhancing FGC realism.
These results establish our method as an effective framework for integrating rich physical priors from physical simulators into generative models, thereby improving the aesthetic appeal of the generated dances when visualized with a human body mesh and advancing their potential for real-world applications. In the future, we aim to further refine both the physical plausibility and smoothness of generated motions.
\begin{table}[t]
\caption{FGD reward and guidance ablation.}
\label{tab:ablation_FGC}
\centering
\footnotesize
\setlength{\tabcolsep}{6pt}
\begin{tabular}{lcc}
\toprule
Method & PFC $\downarrow$ & FGD $\downarrow$ \\
\midrule
EDGE            & 0.8715 & 11.8547 \\
Ours w/o R.\&G. & 0.8728 & 11.1405 \\
Ours w/o G.     & 0.7928 & 10.6357 \\
Ours            & \textbf{0.3361} & \textbf{2.5153} \\
\midrule
Ground Truth    & 1.4699 & 4.9451 \\
\bottomrule
\end{tabular}
\end{table}

\subsection{More metrics~\label{sec:more_metric_maintext}}
To better evaluate the quality of generated motions, we add three more metrics to evaluate the results on the AIST++ dataset. The first metric is MotionCritic Score proposed in~\cite{motioncritic2025}, which uses a pretrained network to rate the motion quality. The second metric is NRDF Score proposed in~\cite{he24nrdf}, which measures the distance of a pose to the manifold of plausible articulated poses. The third metric is FID. However, the feature extractor we use is a pretrained motion autoencoder as in~\cite{zhao2025freedance}. As shown in Table~\ref{tab:newmetric_maintext}, our method achieves the best MotionCritic Score and NRDF Score, while the FID is slightly worse than EDGE, which may be due to the distribution shift during RL-finetuning.
\begin{table}[h]
    \caption{FID, MotionCritic Score, and NRDF Score on AIST++ dataset}
    \label{tab:newmetric_maintext}
    \centering
    \small
    \resizebox{\linewidth}{!}{\begin{tabular}{lccc}
    \toprule
        Method&FID~\cite{zhao2025freedance} $\downarrow$&MotionCritic~\cite{motioncritic2025} $\uparrow$&NRDF~\cite{he24nrdf} $\downarrow$\\
        \midrule
        Ours&0.5420&\textbf{2.26}&\textbf{1.9881}\\
         EDGE&\textbf{0.4988}&-4.10&2.5036\\
         PAMD&0.6193&-5.00&2.1659\\
         \midrule
         FACT&0.5829&-4.87&2.0825\\
         Bailando&0.6072&-9.88&24.2318\\
         Bailando++&0.5866&-2.98&3.2060\\
         DanceBA&0.5563&-3.74&8.8400\\
         \bottomrule
    \end{tabular}}
\end{table}

\subsection{Discussion About the FID~\label{sec:fid_discussion_maintext}}
\begin{figure}[h]
    \centering
    \includegraphics[width=\linewidth]{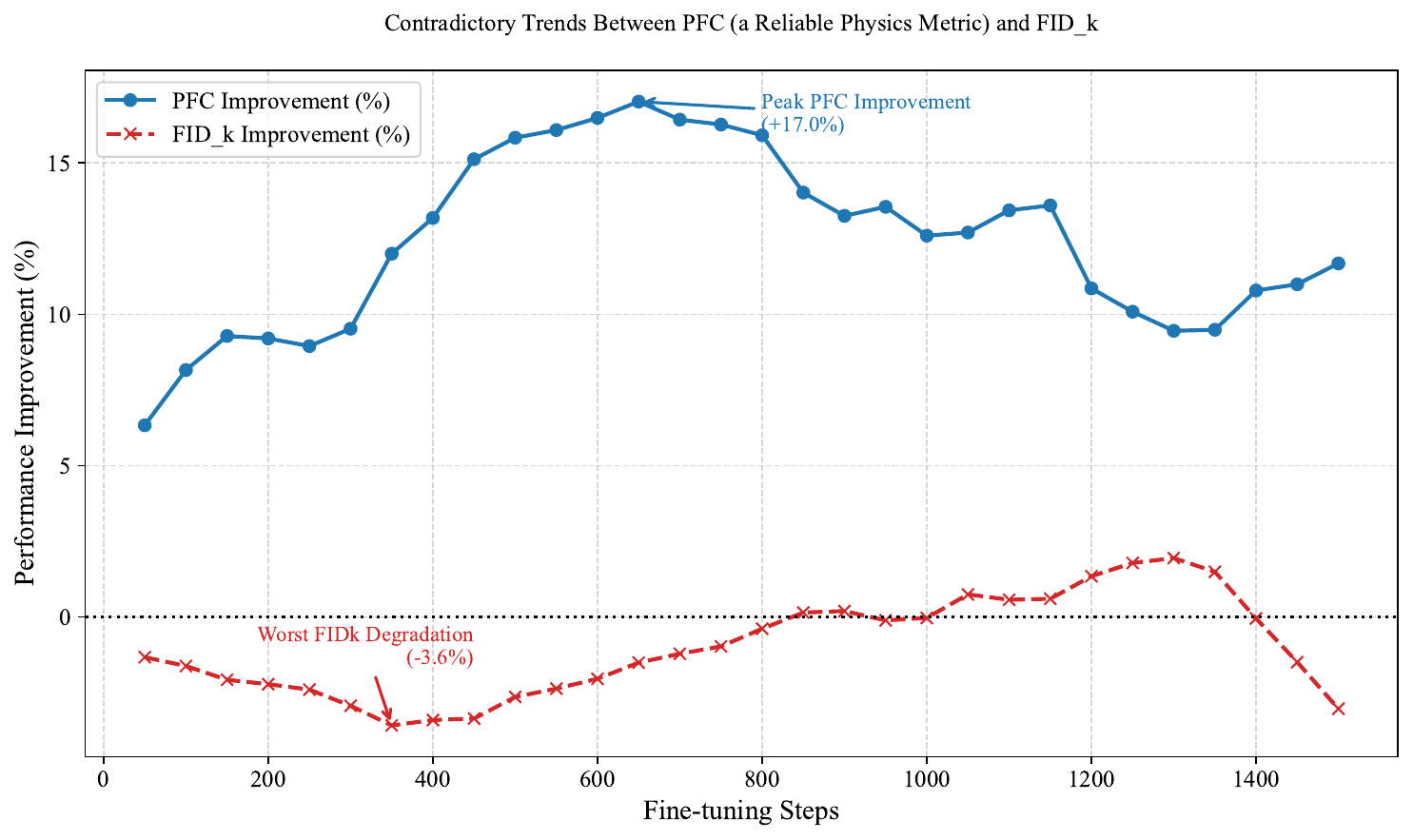}
    \caption{The performance of $\text{FID}_\text{k}$ and PFC during the fine-tuning process of EDGE.}
    \label{fig:pfcvsfid_maintext}
\end{figure}

Many previous works~\cite{tseng2023edge,Luo_2024_CVPR,10264209, huang2024beat,zhang2024bidirectional, wang2025dancechat,motioncritic2025, lin2025quest,zhang2025opendance} have shown that FID is not a good metric to measure the quality of the generated motions in Dance Generation. We also observed that FID is not a reliable metric for motion quality. For instance, when we fine-tuned POPDG with a higher weighted anti-freezing reward, the $\text{FID}_\text{k}/\text{FID}_\text{g}$ score improved to 73.72/29.40 (origin POPDG 76.23/33.37), while the PFC degraded to 2.5438 (origin POPDG 1.9265). This negative correlation underscores FID's unreliability. Moreover, we find that the POPDG training set has a high PFC of 17.2608, as its ground truth motions were generated through motion estimation, which introduces jitter. This causes FID to develop a preference for similarly jittery motions, making it an unreliable metric for dance quality. From Table~\ref{tab:FID_maintext}, the fact that PAMD has worse PFC and better FID compared to EGDE also shows the unreliability of FID. Furthermore, we also plot the change of $\text{FID}_\text{k}$ and $\text{PFC}$ during our fine-tuning process of EDGE. As shown in Fig.~\ref{fig:pfcvsfid_maintext}, the PFC and $\text{FID}_\text{k}$ also exhibit a negative correlation, which further suggests that FID may not be a reliable metric for evaluating dance quality.

\begin{table}[h]
    \caption{FID metrics}
    \label{tab:FID_maintext}
    \centering
    \small
    \begin{tabular}{lc}
    \toprule
        Method & $\text{FID}_\text{k}$ /$\text{FID}_\text{g}$ \\
        \midrule
        EDGE+Ours(20s) &63.59/24.78\\
        EDGE(20s) & 61.67/20.77\\
         \midrule
         EDGE+Ours(5s)&48.18/19.38\\
         EDGE(5s)&47.66/16.23\\
         \midrule
         PAMD&56.90/19.87\\
         \midrule
         FACT&69.04/19.63\\
         Bailando&26.99/10.25\\
         Bailando++&21.71/9.66\\
         DanceBA&31.90/16.22\\
         \bottomrule
    \end{tabular}
\end{table}

Therefore, following~\cite{tseng2023edge, lin2025quest}, we use user study, PFC, music dance alignment, and other physical metrics to measure the motion quality. And we also provide more metrics in Section~\ref{sec:more_metric_maintext} to prove the effectiveness of our method.

\subsection{Advantages Against PhysDiff~\label{sec:phys_maintext}}
\begin{figure}[htbp]
    \centering
    \includegraphics[width=\linewidth]{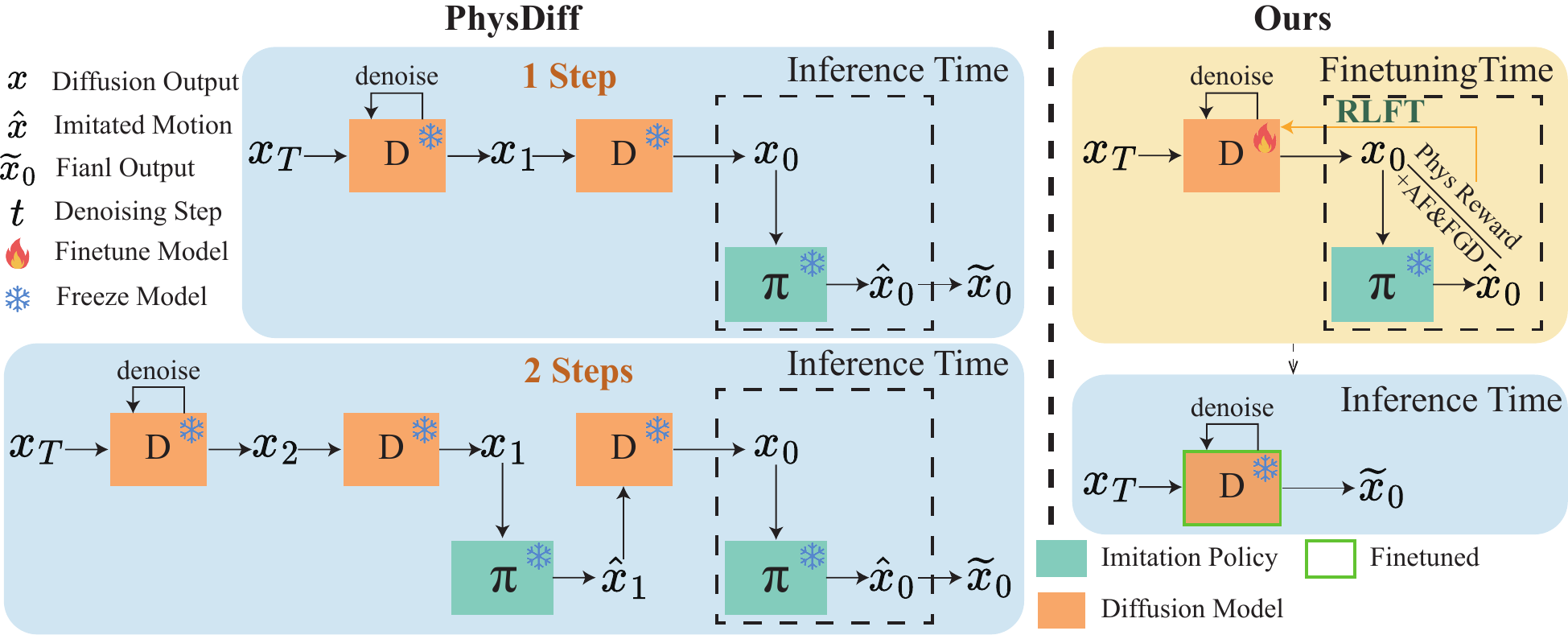}
    \caption{Key differences: PhysDiff vs Ours.}
    \label{fig:physvsour_maintext}
\end{figure}

The key differences between our method and PhysDiff are shown in Fig.~\ref{fig:physvsour_maintext}. Our method instills physical constraints from the simulator into the diffusion model, eliminating the need for an imitation policy during inference.

Thus, the advantages of our method against PhysDiff in dance generation are two folds.

First, in preliminary experiments, we find that though the imitation policy is well-trained on motion datasets, the ability and generalization of imitation policy are limited, which will cause jittering and failure cases. Furthermore, due to the complex nature of dancing motion and the body collision detection introduced in the simulator, the imitation policy becomes more challenging to succeed. As shown in Table~\ref{tab:physdiff_maintext}, the success rate is 95\% for 1 step projection. Moreover, the imitation error will accumulate during the diffusion process. After iterative projection for 4 times (same setting as original PhysDiff), the success rate will decrease significantly to 45\%, which seriously affected the visual quality. That is also why we did not follow the exact settings of PhysDiff. However, we are able to eliminate these artifacts in our proposed method.

Second, the motion projection is time-consuming. As shown in Table~\ref{tab:physdiff_maintext}, our method can save a significant amount of time, making it more computationally efficient.

Moreover, collision detection is disabled in the original paper. In our reproduction, we enable collision checking in the physics simulator to enforce full-body mesh constraints, which is important for bridging the skeleton-to-mesh gap.

\begin{table*}[t]
    \caption{The reproduction of GENMO}
    \label{tab:genmo_reproduction_maintext}
    \centering
    \small
     \begin{tabular}{lcccccc}
    \toprule
        Method&Pen. Rate $\downarrow$&PFC $\downarrow$&FGD $\downarrow$&BAS $\uparrow$&Mot. Mag. $\rightarrow$&$\text{Div}_\text{k}$ /$\text{Div}_\text{g}$ $\rightarrow$\\
         \midrule
         $\text{GENMO}^\diamondsuit$&146.71&0.2904&22.7392&0.2013&0.2374&1.96/5.00\\
        GENMO*&228.04&0.5944&34.9455&\textbf{0.2355}&0.4144&3.08/4.47\\         
         GENMO* w/ Ours&\textbf{138.78}&\textbf{0.1703}&\textbf{5.2832}&0.2309&0.3961&2.81/4.04\\
         \bottomrule
    \end{tabular}

    \vspace{0.5ex}
    \footnotesize
    \raggedright
    $\diamondsuit$: Results from the official checkpoint, which tends to generate freezing dances. *: Results from our re-implementation based on the paper.
\end{table*}

\begin{table}[ht]
    \caption{Advantage anaginst PhysDiff*}
    \label{tab:physdiff_maintext}
    \centering
    \small
    \resizebox{\linewidth}{!}{
    \begin{tabular}{lccc}
    \toprule
        Method &PFC $\downarrow$&Succ. Rate$\uparrow$&Infer. Time$\downarrow$\\
        \midrule
         PhysDiff*(1step)&1.6240&95\%&30s\\
         PhysDiff*(4step)&3.7823&45\%&120s\\
         Ours&\textbf{0.3361}&\textbf{100\%}&\textbf{3s}\\
         \bottomrule
    \end{tabular}}
\end{table}

\subsection{The Reproduction of GENMO~\label{sec:genmo_maintext}}
Although GENMO has released a checkpoint, it appears to be incorrect, and they have not provided detailed training and evaluation guidance. Therefore, we reproduced the model from scratch based on the original paper and applied our proposed method to GENMO*, our reproduction. As shown in Table~\ref{tab:genmo_reproduction_maintext}, the official model ($\text{GENMO}^\diamondsuit$) exhibits better physical plausibility (lower Pen. Rate and PFC) compared to our reproduction (GENMO*), but scores lower on the BAS metric. We attribute this trade-off to the tendency of $\text{GENMO}^\diamondsuit$ to generate freezing motions, proved by its significantly lower Motion Magnitude (Mot. Mag.).

\subsection{Why Dance Generation?}
We will further elaborate on the significance of selecting dance generation as the research focus. First, compared to general motion generation tasks, there are more interactions between different body parts in dance generation, e.g. hand and body contact, causing more frequent self-collision phenomena. This phenomenon severely affects the visual quality of the dance generation when visualized with a human body mesh. Therefore, we think it is more challenging and meaningful to obtain a physically plausible motion in the task of dance generation. Second, in addition to physical plausibility, the dance generation also demands high aesthetic quality. Therefore, we propose an anti-freezing reward specifically for the dance generation, as trivial static movements can diminish the visual quality of the generated dance.

\bibliographystyle{IEEEtran}
\bibliography{main}

\end{document}